\documentclass[12pt]{amsart}

\usepackage{amssymb}
\usepackage{amsmath}
\usepackage{amsthm}
\usepackage{leftidx}
\usepackage{mathdots}
\usepackage{amssymb}
\usepackage{mathrsfs}

\usepackage{epsfig}
\usepackage{verbatim}
\newcommand{\ra}{{\rightarrow}}
\newcommand{\lra}{{\longrightarrow}}
\newcommand{\eproof}{\hfill\rule{2.2mm}{3.0mm}}

\newcommand{\Proof}{\noindent {\bf Proof.~~}}
\newcommand{\D}{{\mathcal D}}

\newcommand{\R}{{\mathbb R}}

\newcommand{\E}{{\mathbb E}}

\newcommand{\supp}{{\rm supp}}
\renewcommand{\eqref}[1]{(\ref{#1})}

\newcommand{\A}{{\mathcal A}}
\newcommand{\X}{{\mathcal X}}
\newcommand{\Y}{{\mathcal Y}}
\newcommand{\B}{{\mathcal B}}

\newcommand{\diag}{{\rm diag}}

\newcommand{\vx}{{\mathbf x}}
\newcommand{\vy}{{\mathbf y}}
\newcommand{\vz}{{\mathbf z}}

\newcommand{\vxi}{{\boldsymbol{\xi}}}

\newcommand{\vmu}{{\boldsymbol{\mu}}}

\newcommand{\vsigma}{{\boldsymbol{\sigma}}}

\newtheorem{prop}{Proposition}[section]
\newtheorem{lem}[prop]{Lemma}
\newtheorem{defi}{Definition}[section]

\newtheorem{theo}[prop]{Theorem}

\begin{document}
\baselineskip 18pt
\title[A Beginner's Guide to GAN]{A Mathematical Introduction to Generative Adversarial Nets (GAN)}
\author{Yang Wang}
\thanks{The author is supported in part by the Hong Kong Research Grant Council grants 16308518 and 16317416, as well as HK Innovation Technology Fund ITS/044/18FX
}
\address{Department of Mathematics  \\ Hong Kong University of Science and Technology\\
Clear Water Bay, Kowloon, Hong Kong}
\email{yangwang@ust.hk}

\begin{abstract}
   Generative Adversarial Nets (GAN) have received considerable attention since the 2014 groundbreaking work by Goodfellow et al \cite{Goodfellow14}. Such attention has led to an explosion in new ideas, techniques and applications of GANs. To better understand GANs we need to understand the mathematical foundation behind them. This paper attempts to provide an overview of GANs from a mathematical point of view. Many students in mathematics may find the papers on GANs more difficulty to fully understand because most of them are written from computer science and engineer point of view. The aim of this paper is to give more mathematically oriented students an introduction to GANs in a language that is more familiar to them. 
\end{abstract}
\maketitle

\section{Introduction}
\setcounter{equation}{0}

\subsection{Background}

Generative Adversarial Nets (GAN) have received considerable attention since the 2014 groundbreaking work by Goodfellow et al \cite{Goodfellow14}. Such attention has led to an explosion in new ideas, techniques and applications of GANs. Yann LeCun has called {\em ``this (GAN) and the variations that are now being proposed is the most interesting idea in the last 10 years in ML, in my opinion.''} In this note I will attempt to provide a beginner's introduction to GAN from a more mathematical point of view, intended for students in mathematics. Of course there is much more to GANs than just the mathematical principle. To fully understand GANs one must also look into their algorithms and applications. Nevertheless I believe that understanding the mathematical principle is a crucial first step towards understanding GANs, and with it the other aspects of GANs will be considerably easier to master.

The original GAN, which we shall refer to as the {\em vanilla GAN} in this paper, was introduced in \cite{Goodfellow14} as a new generative framework from training data sets. Its goal  was to address the following question: Suppose we are given a data set of objects with certain degree of consistency, for example, a collection of images of cats, or handwritten Chinese characters, or Van Gogh painting etc., can we artificially generate similar objects?

This question is quite vague so we need to make it more mathematically specific. We need to clarify what do we mean by {\em ``objects with certain degree of consistency''} or {\em ``similar objects''}, before we can move on.

First we shall assume that our objects are points in $\R^n$. For example, a grayscale digital image of 1 megapixel can be viewed as a point in $\R^n$ with $n = 10^6$. Our data set (training data set) is simply a collection of points in $\R^n$, which we denote by $\X \subset \R^n$. When we say that the objects in the data set $\X$ have certain degree of consistency we mean that they are samples generated from a common  probability distribution $\mu$ on $\R^n$, which is often assumed to have a density function $p(x)$. Of course by assuming $\mu$ to have a density function mathematically we are assuming that $\mu$ is absolutely continuous. Some mathematicians may question the wisdom of this assumption by pointing out that it is possible (in fact even likely) that the objects of interest lie on a lower dimensional manifold, making $\mu$ a singular probability distribution. For example, consider the MNIST data set of handwritten digits. While they are $28 \times 28$ images (so $n= 784$), the actual dimension of these data points may lie on a manifold with much smaller dimension (say the actual dimension may only be 20 or so). This is a valid criticism. Indeed when the actual dimension of the distribution is far smaller than the ambient dimension various problems can arise, such as failure to converge or the so-called {\em mode collapsing}, leading to poor results in some cases. Still, in most applications this assumption does seem to work well. Furthermore, we shall show that the requirement of absolute continuity is not critical to the GAN framework and can in fact be relaxed.

Quantifying ``similar objects'' is a bit trickier and holds the key to GANs. There are many ways in mathematics to measure similarity. For example, we may define a distance function and call two pints $\vx, \vy$ ``similar'' if the distance between them is small. But this idea is not useful here. Our objective is not to generate objects that have small distances to some existing objects in $\X$. Rather we want to generate new objects that may not be so close in whatever distance measure we use to any existing objects in the training data set $\X$, but we feel they belong to the same class. A good analogy is we have a data set of Van Gogh paintings. We do not care to generate a painting that is a perturbation of Van Gogh's Starry Night. Instead we would like to generate a painting that a Van Gogh expert will see as a new Van Gogh painting she has never seen before.

A better angle, at least from the perspective of GANs, is to define similarity in the sense of probability distribution. Two data sets are considered similar if they are samples from the same (or approximately same) probability distribution. Thus more specifically we have our training data set $\X \subset \R^n$ consisting of samples from a probability distribution $\mu$ (with density $p(\vx)$), and we would like to find a probability distribution $\nu$ (with density $q(\vx)$) such that $\nu$ is a good approximation of $\mu$. By taking samples from the distribution $\nu$ we obtain generated objects that are ``similar'' to the objects in $\X$.

One may wonder why don't we just simply set $\nu=\mu$ and take samples from $\mu$. Wouldn't that give us a perfect solution? Indeed --- if we know what $\mu$ is. Unfortunately that is exactly our main problem: we don't know. All we know is a finite set of samples $\X$ drawn from the distribution $\mu$. Hence our real challenge is to learn the distribution $\mu$ from only a finite set of samples drawn over it. We should view finding $\nu$ as the process of approximating $\mu$. GANs do seem to provide a novel and highly effective way for achieving this goal. In general the success of a GAN will depend on the complexity of the distribution $\mu$ and the size of the training data set
$\X$. In some cases the cardinality $|\X|=N$ can be quite large, e.g. for ImageNet data set $N$ is well over $10^7$. But in some other cases, such as Van Gogh paintings, the size $N$ is rather small, in the order of 100 only.

\subsection{The Basic Approach of GAN}

To approximate $\mu$, the vanilla GAN and subsequently other GANs start with an initial probability distribution $\gamma$ defined on $\R^d$, where $d$ may or may not be the same as $n$. For the time being we shall set $\gamma$ to be the standard normal distribution $N(0, I_d)$, although we certainly can choose $\gamma$ to be other distributions. The technique GANs employ is to find a mapping (function) $G:~\R^d \lra \R^n$ such that if a random variable $\vz\in\R^d$ has distribution $\gamma$ then $G(\vz)$ has distribution $\mu$. Note that the distribution of $G(\vz)$ is $\gamma\circ G^{-1}$, where $G^{-1}$ maps subsets of $\R^n$ to subsets of $\R^d$. Thus we are looking for a $G(\vz)$ such that $\gamma\circ G^{-1} = \mu$, or at least is a good approximation of $\mu$. Sounds simple, right?

Actually several key issues remain to be addressed. One issue is that we only have samples from $\mu$, and if we know $G$ we can have samples $G(\vz)$ where $\vz$ is drawn from the distribution $\gamma$. How do we know from these samples that our distribution $\gamma\circ G^{-1}$ is the same or a good approximation of $\mu$? Assuming we have ways to do so, we still have the issue of finding $G(\vz)$.

The approach taken by the vanilla GAN is to form an adversarial system from which $G$ continues to receive updates to improve its performance. More precisely it introduces a ``discriminator function'' $D(x)$, which tries to dismiss the samples generated by $G$ as fakes. The discriminator $D(x)$ is simply a classifier that tries to distinguish samples in the training set $\X$ (real samples) from the generated samples $G(\vz)$ (fake samples). It assigns to each sample $\vx$ a probability $D(\vx) \in [0,1]$ for its likelihood to be from the same distribution as the training samples.  When samples $G(\vz_j)$ are generated by $G$, the discriminator $D$ tries to reject them as fakes. In the beginning this shouldn't be hard because the generator $G$ is not very good. But each time $G$ fails to generate samples to fool $D$, it will learn and adjust with an improvement update. The improved $G$ will perform better, and now it is the discriminator  $D$'s turn to update itself for improvement. Through this adversarial iterative process an equilibrium is eventually reached, so that even with the best discriminator $D$ it can do no better than random guess. At such point, the generated samples should be very similar in distribution to the training samples $\X$.

So one may ask: where do neural networks and deep learning have to do with all this? The answer is that we basically have the fundamental faith that deep neural networks can be used to approximate just about any function, through proper tuning of the network parameters using the training data sets. In particular neural networks excel in classification problems. Not surprisingly, for GAN we shall model both the discriminator function $D$ and the generator function $G$ as neural networks with parameters $\omega$ and $\theta$, respectively. Thus we shall more precisely write $D(x)$ as $D_\omega(x)$ and $G(z)$ as $G_\theta (z)$, and denote $\nu_\theta := \gamma\circ G^{-1}_\theta$. Our objective is to find the desired $G_\theta(\vz)$ by properly tuning $\theta$.

\section{Mathematical Formulation of the Vanilla GAN}
\setcounter{equation}{0}

The adversarial game described in the previous section can be formulated mathematically by minimax of a target function between the discriminator function $D(x): \R^n \lra [0,1]$ and the generator function $G: \R^d \lra\R^n$. The generator $G$ turns random samples $\vz\in\R^d$ from distribution $\gamma$ into generated samples $G(\vz)$. The discriminator $D$ tries to tell them apart from the training samples coming from the distribution $\mu$, while $G$ tries to make the generated samples as similar in distribution to the training samples. In \cite{Goodfellow14} a target loss function is proposed to be
\begin{equation}  \label{eq:V(D,G)}
   V(D, G) := \E_{\vx\sim \mu}[\log D(\vx)]
                           + \E_{\vz\sim \gamma}[\log (1- D(G(\vz)))],
\end{equation}
where $\E$ denotes the expectation with respect to a distribution specified in the subscript. When there is no confusion we may drop the subscript. The vanilla GAN solves the minimax problem
\begin{equation}  \label{eq:GAN-minimax}
   \min_G \max_D \,V(D, G) := \min_G\max_D \,\Bigl(\E_{\vx\sim \mu}[\log D(\vx)]
                           + \E_{\vz\sim \gamma}[\log (1- D(G(\vz)))]\Bigr).
\end{equation}
Intuitively, for a given generator $G$, $\max_D \,V(D, G)$ optimizes the discriminator $D$ to reject generated samples $G(\vz)$ by attempting to assign high values to samples from the distribution $\mu$ and low values to generated samples $G(\vz)$. Conversely, for a given discriminator $D$, $\min_G \,V(D, G)$ optimizes $G$ so that the generated samples $G(\vz)$ will attempt to ``fool'' the discriminator $D$ into assigning high values.

Now set $\vy = G(\vz) \in \R^n$, which has distribution $\nu := \gamma\circ G^{-1}$ as $\vz \in\R^d$ has distribution $\gamma$. We can now rewrite $V(D, G)$ in terms of $D$ and $\nu$ as
\begin{align}
  \tilde V(D, \nu) := V(D, G) &= \E_{\vx\sim \mu}[\log D(\vx)]
                           + \E_{\vz\sim \gamma}[\log (1- D(G(\vz)))] \nonumber  \\
       &= \E_{\vx\sim \mu}[\log D(\vx)] + \E_{\vy\sim \nu}[\log (1- D(\vy))] \nonumber  \\
       &= \int_{\R^n} \log D(x) \, d\mu(x) + \int_{\R^n} \log (1-D(y)) \, d\nu(y). \label{eq:V(D,nu)-int}
\end{align}
The minimax problem \eqref{eq:GAN-minimax} becomes
\begin{equation}  \label{eq:GAN-minimax-int}
   \min_G \max_D \,V(D, G) = \min_G \max_D \,\Bigl(\int_{\R^n} \log D(x) \, d\mu(x) + \int_{\R^n} \log (1-D(y)) \, d\nu(y)\Bigr).
\end{equation}
Assume that $\mu$ has density $p(x)$ and $\nu$ has density function $q(x)$ (which of course can only happen if $d \geq n$). Then
\begin{equation}  \label{eq:V(D,nu)-int2}
   V(D,\nu) = \int_{\R^n} \Bigl(\log D(x)p(x)  + \log (1-D(x)) q(x)\Bigr)\, dx.
\end{equation}
The minimax problem \eqref{eq:GAN-minimax} can now be written as
\begin{equation}  \label{eq:GAN-minimax-int}
   \min_G \max_D \,V(D, G) = \min_G \max_D \,\int_{\R^n} \Bigl(\log D(x)p(x)  + \log (1-D(x)) q(x)\Bigr)\, dx.
\end{equation}
Observe that the above is equivalent to $\min_\nu \max_D \tilde V(D, \nu)$ under the constraint that $\nu = \gamma \circ G^{-1}$ for some $G$. But to better understand the minimax problem it helps to examine $\min_\nu \max_D \tilde V(D, \nu)$ without this constraint. For the case where $\mu,\nu$ have densities \cite{Goodfellow14} has established the following results:

\begin{prop}[\cite{Goodfellow14}]  \label{prop:goodfellow1}
    Given probability distributions $\mu$ and $\nu$ on $\R^n$ with densities $p(x)$ and $q(x)$ respectively,
$$
   \max_D \,V(D, \nu) = \max_D \,\int_{\R^n} \Bigl(\log D(x)p(x)  + \log (1-D(x)) q(x)\Bigr)\, dx
$$
is attained by $D_{p,q}(x) = \frac{p(x)}{p(x)+q(x)}$ for $x \in\supp(\mu) \cup \supp(\nu)$.
\end{prop}

The above proposition leads to

\vspace{2mm}

\begin{theo}[\cite{Goodfellow14}] \label{theo-goodfellow2}
   Let $p(x)$ be a probability density function on $\R^n$. For probability distribution $\nu$ with density function $q(x)$ and $D: \R^n \lra [0,1]$ consider the minimax problem
\begin{equation}  \label{eq:GAN-minimax-uncon}
   \min_\nu \max_D \,\tilde V(D, \nu) = \min_\nu \max_D \,\int_{\R^n} \Bigl(\log D(x)p(x)  + \log (1-D(x)) q(x)\Bigr)\, dx.
\end{equation}
 Then the solution is attained with $q(x)=p(x)$ and $D(x) =1/2$ for all $x\in \supp(p)$.
\end{theo}

Theorem \ref{theo-goodfellow2} says the solution to the minimax problem \eqref{eq:GAN-minimax-uncon} is exactly what we are looking for, under the assumption that the distributions have densities. We discussed earlier that this assumption ignores that the distribution of interest may lie on a lower dimensional manifold and thus without a density function. Fortunately,  the theorem actually holds in the general setting for any distributions. We have:

\vspace{2mm}

\begin{theo}  \label{theo:general-dist}
Let $\mu$ be a given probability distribution on $\R^n$. For probability distribution $\nu$ and function $D: \R^n \lra [0,1]$ consider the minimax problem
\begin{equation}  \label{eq:GAN-minimax-uncon2}
   \min_\nu \max_D \,\tilde V(D, \nu) = \min_\nu \max_D \,\int_{\R^n} \Bigl(\log D(x) \, d\mu(x) +  \log (1-D(x)) \, d\nu(x)\Bigr).
\end{equation}
Then the solution is attained with $\nu=\mu$ and $D(x) =\frac{1}{2}$ $\mu$-almost everywhere.
\end{theo}
\Proof The proof follows directly from Theorem \ref{theo:f-gan} and the discussion in Subsection 3.5, Example 2.
\eproof

Like many minimax problems, one may use the alternating optimization  algorithm to solve  \eqref{eq:GAN-minimax-uncon}, which alternates the updating of $D$ and $q$ (hence $G$). An updating cycle consists of first updating $D$ for a given $q$, and then updating $q$ with the new $D$. This cycle is repeated until we reach an equilibrium. The following is given in \cite{Goodfellow14}:

\begin{prop}[\cite{Goodfellow14}]  \label{prop:alternate}
    If in each cycle, the discriminator $D$ is allowed to reach its optimum given $q(x)$, followed by an update of $q(x)$ so as to improve the minimization criterion
$$
    \min_q \,\int_{\R^n} \Bigl(\log D(x)p(x)  + \log (1-D(x)) q(x)\Bigr)\, dx.
$$
Then $q$ converges to $p$.
\end{prop}

Here I have changed the wording a little bit from the original statement, but have kept its essence intact. From a pure mathematical angle this proposition is not rigorous. However, it provides a practical framework for solving the vanilla GAN minimax problem, namely in each cycle we may first optimize the discriminator $D(x)$ all the way for the current $q(x)$, and then update $q(x)$ given the new $D(x)$ just a little bit. Repeating this cycle will lead us to the desire solution. In practice, however, we rarely optimize $D$ all the way for a given $G$; instead we usually update $D$ a little bit before switching to updating $G$. 

Note that the unconstrained minimax problem \eqref{eq:GAN-minimax-uncon} and \eqref{eq:GAN-minimax-uncon2} are not the same as the original minimax problem \eqref{eq:GAN-minimax} or the equivalent formulation \eqref{eq:V(D,nu)-int}, where $\nu$ is constrained to be of the form $\nu = \gamma \circ G^{-1}$. Nevertheless it is reasonable in practice to assume that \eqref{eq:GAN-minimax} and \eqref{eq:V(D,nu)-int} will exhibit similar properties as those being shown in Theorem \ref{theo:general-dist} and Proposition \ref{prop:alternate}. In fact, we shall assume the same even after we further restrict the discriminator and generator functions to be neural networks  $D= D_\omega$ and $G= G_\theta$ as intended. Set $\nu_\theta = \gamma \circ G_\theta^{-1}$. Under this model our minimax problem has become $\min_\theta \max_\omega \,V(D_\omega, G_\theta)$ where
\begin{align}
   V(D_\omega, G_\theta) &=
            \E_{\vx\sim \mu}[\log D_\omega(\vx)]
                  + \E_{\vz\sim \gamma}[\log (1- D_\omega(G_\theta (\vz)))]  \label{eq:GAN-NN}\\
           &=  \int_{\R^n} \Bigl(\log D_\omega(x) \, d\mu(x)
                              + \log (1-D_\omega(x)) \, d\nu_\theta(x)\Bigr). \label{eq:GAN-NN-int}
\end{align}

Equation \eqref{eq:GAN-NN} is the key to carrying out the actual optimization: since we do not have the explicit expression for the target distribution $\mu$, we shall approximate the expectations through sample averages. Thus \eqref{eq:GAN-NN} allows us to approximate $V(D_\omega, G_\theta)$ using samples. More specifically, let $\A$ be a subset of samples from the training data set $\X$ (a minibatch) and $\B$ be a minibatch of samples in $\R^d$ drawn from the distribution $\gamma$. Then we do the approximation
\begin{align}
      \E_{\vx \sim \mu} [\log D_\omega(\vx)] & \approx \frac{1}{|\A|}\sum_{\vx\in\A} \log D_\omega(\vx) \\
      \E_{\vz\sim \gamma}[\log (1- D_\omega(G_\theta (\vz)))] & \approx \frac{1}{|\B|}\sum_{\vz\in\B} \log (1- D_\omega(G_\theta (\vz))).
\end{align}
The following algorithm for the vanilla GAN was presented in \cite{Goodfellow14}:


\par\noindent\rule{\textwidth}{1.5pt}
{\bf Vanilla GAN Algorithm} Minibatch stochastic gradient descent training of generative adversarial nets. The number of
steps to apply to the discriminator, $k$, is a hyperparameter. $k = 1$, the least expensive option,  was used in the experiments in \cite{Goodfellow14}.\\
\noindent\rule{\textwidth}{0.4pt}

\noindent
for number of training iterations do \hfill\\
\hphantom{ab} for k steps do
\begin{itemize}
   \item Sample minibatch of $m$  samples $\{\vz_1, \dots, \vz_m\}$ in $\R^d$ from the distribution $\gamma$.
   \item Sample minibatch of $m$  samples $\{\vx_1, \dots, \vx_m\} \subset \X$ from the training set $\X$.
   \item Update the discriminator $D_\omega$ by ascending its stochastic gradient with respect to $\omega$:
    $$
         \nabla_\omega \frac{1}{m} \sum_{i=1}^m \Bigl[\log D_\omega(\vx_i) +\log (1- D_\omega(G_\theta (\vz_i)))\Bigr].
    $$
\end{itemize}
\hphantom{ab} end for
\begin{itemize}
   \item Sample minibatch of $m$  samples $\{\vz_1, \dots, \vz_m\}$ in $\R^d$ from the distribution $\gamma$.
   \item Update the generator $G_\theta$ by descending its stochastic gradient with respect to $\theta$:
    $$
         \nabla_\theta \frac{1}{m} \sum_{i=1}^m \log (1- D_\omega(G_\theta (\vz_i))).
    $$
\end{itemize}
end for

The gradient-based updates can use any standard gradient-based learning rule. The paper used momentum in their experiments. \\
\noindent\rule{\textwidth}{0.4pt}


Proposition \ref{prop:alternate} serves as a heuristic justification for the convergence of the algorithm. One problem often encountered with the Vanilla GAN Algorithm is that the updating of $G_\theta$ from minimization of $\E_{\vz\sim \gamma}[\log (1- D_\omega(G_\theta (\vz)))]$ may saturate early. So instead the authors substituted it with minimizing $-\E[\log D_\omega(G_\theta(\vz))]$. This is the well-known {\em ``\,$\log D$ trick''}, and it seems to offer superior performance. We shall examine this more closely later on.

\section{$f$-Divergence and $f$-GAN}
\setcounter{equation}{0}

Recall that the motivating problem for GAN is that we have a probability distribution $\mu$ known only in the form of a finite set of samples (training samples). We would like to learn this target distribution through iterative improvement. Starting with a probability distribution $\nu$ we iteratively update $\nu$ so it gets closer and closer to the target distribution $\mu$. Of course to do so we will first need a way to measure the discrepancy between two probability distributions. The vanilla GAN has employed a discriminator for this purpose. But there are other ways.

\subsection{$f$-Divergence}
One way to measure the discrepancy between two probability distributions $\mu$ and $\nu$ is through the  {\em Kullback-Leibler divergence}, or {\em KL divergence}. Let $p(x)$ and $q(x)$ be two probability density functions defined on $\R^n$. The KL-divergence of $p$ and $q$ is defined as
$$
     D_{KL} (p\| q) := \int_{\R^n} \log\Bigl(\frac{p(x)}{q(x)}\Bigr) p(x)\,dx.
$$
Note that $D_{KL} (p\| q) $ is finite only if  $q(x) \neq 0$ on $\supp(p)$ almost everywhere. While KL-divergence is widely used, there are other divergences such as the Jensen-Shannon divergence
$$
     D_{JS}(p\| q) := \frac{1}{2} D_{KL}(p\| M) + \frac{1}{2} D_{KL}(q\| M),
$$
where $M := \frac{p(x)+q(x)}{2}$. One advantage of the Jensen-Shannon divergence is that it is well defined for any probability density functions $p(x)$ and $q(x)$, and is symmetric $D_{JS}(p\| q) = D_{JS}(q \| p)$. In fact, following from Proposition \ref{prop:goodfellow1} the minimization part of the minimax problem in the vanilla GAN is precisely the minimization over $q$ of $D_{JS}(p\| q)$ for a given density function $p$. As it turns out, both $D_{KL}$ and $D_{JS}$ are special cases of the more general {\em $f$-divergence}, introduced by Ali and Silvey \cite{AliSil65}.

Let $f(x)$ be a strictly convex function with domain $I\subseteq \R$ such that $f(1)=0$. Throughout this paper we shall adopt the convention that $f(\vx) = +\infty$ for all $\vx \not\in I$.
\begin{defi}  \label{def:f-divergence}
  Let $p(x)$ and $q(x)$ be two probability density functions  on $\R^n$. Then the {\em $f$-divergence} of $p$ and $q$ is defined as
\begin{equation}  \label{eq:f-divergence}
    D_f (p\| q) := \E_{\vx \sim q}\Big[f\Bigl(\frac{p(\vx)}{q(\vx)}\Bigr)\Bigr] =\int_{\R^n} f\Bigl(\frac{p(x)}{q(x)}\Bigr) q(x)\,dx,
\end{equation}
where we adopt the convention that $f(\frac{p(x)}{q(x)}) q(x) =0$ if $q(x)=0$.
\end{defi}

\noindent
{\bf Remark.}~ Because the $f$-divergence is not symmetric in the sense that $D_f (p\| q) \neq D_f (q\| p)$ in general, there might be some confusion as to which divides which in the fraction. If we follow the original Ali and Silvey paper \cite{AliSil65} then the definition of $D_f (p\| q)$ would be our $D_f (q\| p)$. Here we adopt the same definition as in the paper \cite{NoCsTo16}, which first introduced the concept of $f$-GAN.

\begin{prop}  \label{prop:f-div}
	Let $f(x)$ be a strictly convex function on domain $I\subseteq \R$ such that $f(1)=0$. Assume either $\supp(p) \subseteq \supp(q)$ (equivalent to $p\ll q$) or $f(t)>0$ for $t \in [0,1)$. Then
	$D_f (p\| q) \geq 0$, and $D_f (p\| q)=0$ if and only if $p(x)=q(x)$.
\end{prop}
\Proof  By the convexity of $f$ and Jensen's Inequality
$$
    D_f (p\| q) = \E_{\vx\sim q} \Bigl[f\Bigl(\frac{p(\vx)}{q(\vx)}\Bigr)\Bigr]
              \geq f\Bigl(\E_{\vx\sim q}\Bigl[\frac{p(\vx)}{q(\vx)}\Bigr]\Bigr)
              = f\Bigl(\int_{\supp(q)} p(x)\,dx\Bigr) =: f(r),
$$
where the equality holds if and only if $q(x)/p(x)$ is a constant or $f$ is linear on the range of $p(x)/q(x)$. Since $f$ is strictly convex, it can only be the former. Thus for the equality to hold we must have $p(x) = rq(x)$ on $\supp(q)$.

Now clearly $r \leq 1$. If $\supp(p) \subseteq \supp(q)$ then $r=1$, and we have $D_f (p\| q) \geq 0$. The equality holds if and only if $p=q$. If $f(t)>0$ for all $t \in[0,1)$ then we also have $D_f (p\| q) \geq f(r)\geq 0$. For $r<1$ we will have $D_f (p\| q) \geq f(r)> 0$. Thus if $D_f (p\| q) = 0$ we must have $r=1$ and $p=q$.
\eproof

It should be noted that $f$-divergence can be defined for two arbitrary probability measures $\mu$ and $\nu$ on a probability space $\Omega$. Let $\tau$ be another probability measure such that $\mu, \nu \ll \tau$, namely both $\mu, \nu$ are absolutely continuous with respect to $\tau$. For example, we may take $\tau = \frac{1}{2}(\mu+\nu)$. Let $p = d\mu/d\tau$ and $q = d\nu/d\tau$ be their Radon-Nikodym derivatives. We define the $f$-divergence of $\mu$ and $\nu$ as
\begin{equation}  \label{eq:f-div-singular}
   D_f (\mu\| \nu) := \int_{\Omega} f\Bigl(\frac{p(x)}{q(x)}\Bigr) q(x)\,d\tau
                    = \E_{\vx\sim \nu} \Bigl[f\Bigl(\frac{p(\vx)}{q(\vx)}\Bigr)\Bigr].
\end{equation}
Again we adopt the convention that $f(\frac{p(x)}{q(x)}) q(x) =0$ if $q(x)=0$. It is not hard to show that this definition is independent of the choice for the probability measure $\tau$, and Proposition \ref{prop:f-div} holds for the more general $D_f (\mu\| \nu)$ as well.

With $f$-divergence measuring the discrepancy between two measures, we can now consider applying it to GANs. The biggest challenge here is that we don't have an explicit expression for the target distribution $\mu$. As with the vanilla GAN, to compute $D_f(p\|q)$ we must express it in terms of sample averages. Fortunately earlier work by Nguyen, Wainwright and Jordan \cite{NgWaJo10} has already tackled this problem using the convex conjugate of a convex function.

\subsection{Convex Conjugate of a Convex Function}

The {\em convex conjugate} of a convex function $f(x)$ is also known as the {\em Fenchel transform} or  {\em Fenchel-Legendre transform} of $f$, which is a generalization of the well known Legendre transform. Let $f(x)$ be a convex function defined on an interval $I \subseteq \R$. Then its convex conjugate $f^*: \R \lra \R \cup\{\pm \infty\}$ is defined to be
\begin{equation}  \label{eq:convex-conj}
   f^*(y) = \sup_{t\in I} \,\{ ty - f(t)\}.
\end{equation}
As mentioned earlier we extend $f^*$ to the whole real line by adopting the convention that $f(x)=+\infty$ for $x \not\in I$. Below is a more explicit expression for $f^*(y)$.
\begin{lem}  \label{lem:f*formula}
   Assume that $f(x)$ is strictly convex and continuously differentiable on its domain $I \subseteq \R$, where  $I^o =(a,b)$ with $a,b \in [-\infty, +\infty]$.   Then
   \begin{equation} \label{eq:f*formula}
       f^*(y) = \left\{\begin{array}{ll}   y f'^{-1}(y) - f(f'^{-1}(y)),\phantom{aa
       }  &  y\in f'(I^o) \\
                       \lim_{t\ra b^-} (ty -f(t)),     & y \geq \lim_{t\ra b^-} f'(t) \\
                       \lim_{t\ra a^+} (ty -f(t)),     & y \leq \lim_{t\ra a^+} f'(t).
                         \end{array}
               \right.
   \end{equation}
\end{lem}
\Proof  Let $g(t) = ty -f(t)$. The $g'(t) = y - f'(t)$ on $I$, which is strictly decreasing by the convexity of $f(t)$. Hence $g(t)$ is strictly concave on $I$. If $y=f'(t^*)$ for some $t^*\in I^o$ then $t^*$ is a critical point of $g$ so it must be its global maximum.  Thus $g(t)$ attains its maximum at $t=t^* = f'^{-1}(y)$. Now assume $y$ is not in the range of $f'$ then $g'(t)>0$ or $g'(t)<0$ on $I^o$. Consider the case  $g'(t)>0$ for all $t\in I^o$. Clearly the supreme of $g(t)$ is achieved as $t \ra b^-$ since $g(t)$ is monotonously increasing. The case for $g'(t)<0$ for all $t\in I^o$ is similarly derived.
\eproof

\vspace{2mm}
\noindent
{\bf Remark.}~~Note that $+\infty$ is a possible value for $f^*$. The domain ${\rm Dom}\,(f^*)$ for $f^*$ is defined as the set on which $f^*$ is finite.

A consequence of Lemma \ref{lem:f*formula} is that under the assumption that $f$ is continuously differentiable, $\sup_{t\in I} \,\{ ty - f(t)\}$ is attained for some $t\in I$ if and only if $y$ is in the range of $f'(t)$. This is clear if $y \in f'(I^o)$, but it can also be argued rather easily for finite boundary points of $I$. More generally without the assumption of differentiability, $\sup_{t\in I} \,\{ ty - f(t)\}$ is attained if and only if $y \in \partial f(t)$ for some $t\in I$, where $\partial f(t)$ is the set of sub-derivatives. The following proposition summarizes some important properties of convex conjugate:

\begin{prop}   \label{prop:convex-dual}
    Let $f(x)$ be a convex function on $\R$ with range in $\R \cup\{\pm\infty\}$. Then $f^*$ is convex and is lower-semi continuous. Furthermore if $f$ is  lower-semi continuous then it satisfies the Fenchel Duality $f = (f^*)^*$.
\end{prop}
\Proof This is a well known result. We omit the proof here.
\eproof

The table below lists the convex dual of some common convex functions:

\begin{center}
\begin{tabular}{ |c|c| }
 \hline
 $f(x)$ & $f^*(y)$  \\
 \hline
 $\displaystyle f(x)=-\ln(x),~~x>0$ & $\displaystyle f^*(y)=-1 - \ln(-y)),~~y<0$ \\
 \hline
 $\displaystyle f(x)=e^x$ & $\displaystyle f^*(y)=y\ln(y)-y,~~y>0$ \\
 \hline
 $\displaystyle f(x)=x^2$ & $\displaystyle f^*(y)= \frac{1}{4}y^2$ \\
 \hline
 $\displaystyle f(x)=\sqrt{1+x^2}$ & $\displaystyle f^*(y)= -\sqrt{1-y^2},~~y\in [-1,1]$ \\
 \hline
 $\displaystyle f(x)=0,~~x\in[0,1]$ & $\displaystyle f^*(y)= \mbox{ReLu}(y)$ \\
 \hline
 $\displaystyle f(x)=g(ax-b),~~a\neq 0$ & $\displaystyle f^*(y)= \frac{b}{a}y +g^*(\frac{y}{a})$ \\
 \hline
\end{tabular}
\end{center}

\subsection{Estimating $f$-Divergence Using Convex Dual}

To estimate $f$-divergence from samples, Nguyen, Wainwright and Jordan \cite{NgWaJo10} has proposed the use of the convex dual of $f$. Let $\mu,\nu$ be probability measures such that $\mu,\nu \ll \tau$ for some probability measure $\tau$, with $p = d\mu/d\tau$ and $q = d\nu/d\tau$.  In the nice case of $\mu \ll \nu$, by $f(x) =(f^*)^*(x)$ we have
\begin{align}
     D_f (\mu\| \nu) & := \int_{\Omega} f\Bigl(\frac{p(x)}{q(x)}\Bigr) q(x)\,d\tau \nonumber\\
                    & =\int_{\Omega}\sup_{t} \Bigl\{t\,\frac{p(x)}{q(x)} - f^*(t)\Bigr\} q(x)\,d\tau (x)
                                                                    \label{eq:D_f-sup1}\\
                    & = \int_{\Omega}\sup_{t}\, \Bigl\{t p(x) - f^*(t) q(x)\Bigr\} \,d\tau (x)
                                                                     \label{eq:D_f-sup2}\\
                    & \geq \int_{\Omega} \Bigl(T(x) p(x) - f^*(T(x)) q(x)\Bigr) \,d\tau (x) \nonumber\\
                    & = \E_{\vx\sim \mu} [T(\vx)] -\E_{\vx\sim \nu} [f^*(T(\vx))] \nonumber
\end{align}
where $T(x)$ is any Borel function. Thus taking $T$ over all Borel functions we have

\begin{equation}  \label{eq:D_f-upperbound}
     D_f (\mu\| \nu) \geq \sup_{T}  \Bigl(\E_{\vx\sim \mu} [T(\vx)] -\E_{\vx\sim \nu} [f^*(T(\vx))]\Bigr).
\end{equation}
On the other hand,  note that for each $x$, $\sup_{t} \bigl\{t\frac{p(x)}{q(x)} - f^*(t)\bigr\}$ is attained for some $t=T^*(x)$ as long as $\frac{p(x)}{q(x)}$ is in the range of sub-derivatives of $f^*$. Thus if this holds for all $x$ we have
$$
     D_f (\mu\| \nu) = \Bigl(\E_{\vx\sim \nu} [T^*(\vx)] -\E_{\vx\sim \mu} [f^*(T^*(\vx))]\Bigr).
$$
In fact, equality holds in general under mild conditions.

\begin{theo}    \label{theo:sample-f-div}
Let $f(t)$ be strictly convex and continuously differentiable on $I \subseteq \R$. Let $\mu,\nu$ be Borel probability measures on $\R^n$ such that $\mu \ll \nu$. Then
\begin{equation}   \label{eq:f-div-equality}
     D_f (\mu\| \nu) = \sup_{T}  \Bigl(\E_{\vx\sim \mu} [T(\vx)] -\E_{\vx\sim \nu} [f^*(T(\vx))]\Bigr),
\end{equation}
where $\sup_T$ is taken over all Borel functions $T:~\R^n \lra \,{\rm Dom}\,(f^*)$. Furthermore assume that 
$p(x) \in I$ for all $x$. Then $T^*(x) := f'(p(x))$ is an optimizer of \eqref{eq:f-div-equality}.
\end{theo}
\Proof We have already establish the upper bound part \eqref{eq:D_f-upperbound}. Now we establish the lower bound part. Let $p(x) = d\mu/d\nu(x)$. We examine \eqref{eq:D_f-sup1} with $q(x)=1$ and $\sup_ t\,\{t p(x)-f^*(t)\}$ for each $x$. Denote $g_x(t) = t p(x)-f^*(t)$. Let $S= {\rm Dom}\,(f^*)$ and assume $S^o =(a,b)$ where $a,b \in \R\cup\{\pm\infty\}$. We now construct a sequence $T_k(x)$ as follows: If $p(x)$ is in the range of ${f^*}'$, say $p(x) = {f^*}'(t_x)$, we set $T_k(x) = t_x \in S$. If $p(x)-{f^*}'(t)>0$ for all $t$ then $g_x(t)$ is strictly increasing. The supreme of $g_x(t)$ is attained at the boundary point $b$, and we will set $T_k(x) = b_k \in S$ where $b_k \ra b^-$. If $p(x)-{f^*}'(t)<0$ for all $t$ then $g_x(t)$ is strictly decreasing. The supreme of $g_x(t)$ is attained at the boundary point $a$, and we will set $T_k(x) = a_k \in S$ where $a_k \ra a^+$. By Lemma \ref{lem:f*formula} and its proof we know that
$$
    \lim_{k\ra\infty} \Bigl(T_k(x)p(x) - f^*(T_k(x))\Bigr) =
       \sup_{t} \,\Bigl\{tp(x) - f^*(t)\Bigr\}.
$$
Thus
$$
    \lim_{k\ra\infty}\Bigl(\E_{\vx\sim \nu} [T_k(\vx)] -\E_{\vx\sim \mu} [f^*(T_k(\vx))]\Bigr) = D_f (\mu\| \nu),
$$

To establish the last part of the theorem, assume that $p(x) \in I$. By Lemma \ref{lem:f*formula}, set $s(t) = f'^{-1}(t)$ for $t$ in the range of $f'$ so we have
$$
     {f^*}'(t) = \Bigl(ts(t)-f(s(t))\Bigr)' = s(t)+ts'(t)-f'(s(t))s'(t) = s(t).
$$
Thus $g_x'(t) = p(x)-{f^*}'(t) = p(x) - f'^{-1}(t)$. It follows that $g_x(t)$ attains its maximum at $t = f'(p(x))$. This proves that $T^*(x) = f'(p(x))$ is an optimizer for \eqref{eq:f-div-equality}.
\eproof

The above theorem requires that $\mu \ll \nu$. What if this does not hold? We have

\begin{theo}    \label{theo:sample-f-div-singular}
Let $f(t)$ be convex such that the domain of $f^*$ contains $(a,\infty)$ for some $a\in\R$. Let $\mu,\nu$ be Borel probability measures on $\R^n$ such that $\mu \not\ll \nu$. Then
\begin{equation}   \label{eq:f-div-singular}
     \sup_{T}  \Bigl(\E_{\vx\sim \mu} [T(\vx)] -\E_{\vx\sim \nu} [f^*(T(\vx))]\Bigr) =+\infty,
\end{equation}
where $\sup_T$ is taken over all Borel functions $T:~\R^n \lra \,{\rm Dom}\,(f^*)$.
\end{theo}
\Proof Take $\tau = \frac{1}{2}(\mu+\nu)$. Then $\mu,\nu\ll \tau$. Let $p = d\mu/d\tau$ and $q = d\nu/d\tau$ be their Radon-Nikodym derivatives. Since $\mu \not\ll \nu$ there exists a set $S_0$ with $\mu(S_0)>0$ on which $q(x)=0$. Fix a $t_0$ in the domain of $f^*$. Let $T_k(x)= k$ for $x\in S_0$ and $T_k(x) = t_0$ otherwise. Then
$$
    \E_{\vx\sim \mu} [T_k(\vx)] -\E_{\vx\sim \nu} [f^*(T_k(\vx))]
                    \geq k\mu(S_0) - f^*(t_0) (1-\nu(S_0)) \lra +\infty.
$$
This proves the theorem.
\eproof

As one can see, we clearly have a problem in the above case. If the domain of $f^*$ is not bounded from above, \eqref{eq:f-div-equality} does not hold unless $\mu \ll \nu$. In many practical applications the target distribution $\mu$ might be singular, as the training data we are given may lie on a lower dimensional manifold. Fortunately there is still hope as given by the next theorem:

\begin{theo}    \label{theo:sample-f-div-singular-2}
Let $f(t)$ be a lower semi-continuous convex function such that the domain $I^*$ of $f^*$ has $\sup I^* = b^* <+\infty$. Let $\mu,\nu$ be Borel probability measures on $\R^n$ such that $\mu = \mu_s +\mu_{ab}$, where $\mu_s \perp \nu$ and $\mu_{ab} \ll \nu$. Then
\begin{equation}   \label{eq:f-div-singular-2}
     \sup_{T}  \Bigl(\E_{\vx\sim \mu} [T(\vx)] -\E_{\vx\sim \nu} [f^*(T(\vx))]\Bigr) =D_f(\mu\|\nu) + b^* \mu_s(\R^n),
\end{equation}
where $\sup_T$ is taken over all Borel functions $T:~\R^n \lra \,{\rm Dom}\,(f^*)$.
\end{theo}
\Proof Again, take $\tau = \frac{1}{2}(\mu+\nu)$. Then $\mu,\nu\ll \tau$. The decomposition $\mu = \mu_{ab} + \mu_s$ where $\mu_{ab} \ll \nu$ and $\mu_s \perp \nu$ is unique and guaranteed by the Lebesgue Decomposition Theorem. Let $p_{ab} = d\mu_{ab}/d\tau$, $p_{s} = d\mu_{s}/d\tau$ and $q = d\nu/d\tau$ be their Radon-Nikodym derivatives. Since $\mu_s \perp \nu$, we may divide $\R^n$ into $\R^n = \Omega \cup \Omega^c$ where $\Omega =\supp(q)$. Clearly we have $q(x)=p_{ab}(x)=0$ for $x\in \Omega^c$.  Thus
\begin{align*}
    \sup_T\Bigl(&\E_{\vx\sim \mu} [T(\vx)] -\E_{\vx\sim \nu} [f^*(T_k(\vx))]\Bigr) \\
        &=\sup_T \int_{\Omega} \Bigl(T(x) p_{ab}(x)-f^*(T(x))q(x)\Bigr)\,d\tau
             + \sup_T \int_{\Omega^c} T(x)p_{ab}(x)\,d\tau \\
   & =\sup_T \int_{\Omega} \Bigl(T(x)\frac{p_{ab}(x)}{q(x)}-f^*(T(x))\Bigr)q(x)\,d\tau + b^*\mu_s(\Omega^c) \\
   & =\int_{\Omega} f\Bigl(\frac{p_{ab}(x)}{q(x)}\Bigr)q(x)\,d\tau + b^*\mu_s(\R^n) \\
    & =\int_{\Omega} f\Bigl(\frac{p(x)}{q(x)}\Bigr)q(x)\,d\tau + b^*\mu_s(\R^n) \\
    & = D_f(\mu\|\nu) + b^*\mu_s(\R^n).
\end{align*}
This proves the theorem.
\eproof

\subsection{$f$-GAN: Variational Divergence Minimization (VDM)}

We can formulate a generalization of the vanilla GAN using $f$-divergence. For a given probability distribution $\mu$, the {\em $f$-GAN} objective is to minimize the $f$-divergence $D_f(\mu\| \nu)$ with respect to the probability distribution $\nu$. Carried out in the sample space, $f$-GAN solves the following  minimax problem
\begin{equation}   \label{eq:f-GAN}
      \min_\nu \sup_{T} \, \Bigl(\E_{\vx\sim \nu} [T(\vx)] -\E_{\vx\sim \mu} [f^*(T(\vx))]\Bigr).
\end{equation}
The $f$-GAN framework is first introduced in \cite{NoCsTo16}, and the optimization problem \eqref{eq:f-GAN} is referred to as the {\em Variational Divergence Minimization (VDM)}. Note VDM looks similar to the minimax problem in vanilla GAN. The Borel function $T$ here is called a {\em critic function}, or just a {\em critic}. Under the assumption of $\mu\ll\nu$, by Theorem \ref{theo:sample-f-div} this is equivalent to $\min_\nu \,D_f (\mu\| \nu) $. One potential problem of $f$-GAN is that by Theorem \ref{theo:sample-f-div-singular} if $\mu \not\ll \nu$ then \eqref{eq:f-GAN} is in general not equivalent to $\min_\nu \,D_f (\mu\| \nu) $. Fortunately for specially chosen $f$ this is not a problem.

\begin{theo}    \label{theo:f-gan}
	Let $f(t)$ be a lower semi-continuous strictly convex function such that the domain $I^*$ of $f^*$ has $\sup I^* = b^* \in [0,\infty)$. Assume further that $f$ is continuously differentiable on its domain and $f(t)>0$ for $t\in (0,1)$. Let $\mu$ be Borel probability measures on $\R^n$. Then $\nu =\mu$ is the unique optimizer of
	$$
	   \min_\nu \sup_{T} \, \Bigl(\E_{\vx\sim \nu} [T(\vx)] -\E_{\vx\sim \mu} [f^*(T(\vx))]\Bigr),
	$$
	where $\sup_T$ is taken over all Borel functions $T:~\R^n \lra \,{\rm Dom}\,(f^*)$ and $\inf_\nu$ is taken over all Borel probability measures.
\end{theo}
\Proof By Theorem\ref{theo:sample-f-div-singular-2}, for any Borel probability measure $\nu$ we have
$$
   \sup_{T}  \Bigl(\E_{\vx\sim \mu} [T(\vx)] -\E_{\vx\sim \nu} [f^*(T(\vx))]\Bigr) =D_f(\mu\|\nu) + b^* \mu_s(\R^n) \geq D_f(\mu\|\nu).
$$
Now by Proposition \ref{prop:f-div}, $D_f(\mu\|\nu) \geq 0$, and equality holds if and only $\nu=\mu$. Thus $\nu =\mu$ is the unique optimizer.
\eproof

\subsection{Examples}

We shall now look at some examples of $f$-GAN for different choices of the convex function $f$.

\noindent\vspace{3mm}
{\bf Example 1:  \boldmath{$f(t) = -\ln(t)$}}.

This is the KL-divergence. We have $f^*(u) = -1-\ln(-u)$ with domain $I^* = (-\infty, 0)$.  $f$ satisfies all conditions of Theorem \ref{theo:f-gan}. The corresponding $f$-GAN objective is
\begin{equation}   \label{eq:f-GAN-KL}
\min_\nu \sup_{T} \, \Bigl(\E_{\vx\sim \nu} [T(\vx)] +\E_{\vx\sim \mu} [\ln(-T(\vx))]\Bigr)+1,
\end{equation}
where $T(x)<0$. If we ignore the constant $+1$ term and set $D(x) = -T(x)$ then we obtain the equivalent minimax problem
$$
   \min_\nu \sup_{D> 0} \, \Bigl(\E_{\vx\sim \nu} [-D(\vx)] +\E_{\vx\sim \mu} [\ln(D(\vx))]\Bigr).
$$

\noindent\vspace{3mm}
{\bf Example 2:  \boldmath{$f(t) = -\ln(t+1)+\ln(t) +(t+1)\ln2$}}.

This is the Jensen-Shannon divergence. We have $f^*(u) = -\ln(2-e^u)$ with domain $I^* = (-\infty, \ln2)$. Again $f$ satisfies all conditions of Theorem \ref{theo:f-gan}. The corresponding $f$-GAN objective is
\begin{equation}   \label{eq:f-GAN-JS}
\min_\nu \sup_{T} \, \Bigl(\E_{\vx\sim \nu} [T(\vx)] +\E_{\vx\sim \mu} [\ln(2-e^{T(\vx)})]\Bigr),
\end{equation}
where $T(x)<\ln2$. Set $D(x) = 1-\frac{1}{2}e^{T(\vx)}$, and so $T(x) = \ln(1-D(x)) + \ln2$. Substituting in \eqref{eq:f-GAN-JS} yields
$$
      \min_\nu \max_{D>0}  \, \Bigl(\E_{\vx\sim \mu} [\ln(D(\vx))] +\E_{\vx\sim \nu} [\ln(1-D(\vx))]\Bigr) + \ln4.
$$
Ignoring the constant $\ln4$, the vanilla GAN is a special case of $f$-GAN with $f$ being the Jensen-Shannon divergence.

\noindent\vspace{3mm}
{\bf Example 3:  \boldmath{$f(t) = \alpha^{-1} |t-1|$} where  \boldmath{$\alpha>0$} }.

Here we have $f^*(u) = u$ with domain $I^*=[-\alpha, \alpha]$. While $f$ is not strictly convex and continuously differentiable, it does satisfy the two important conditions of Theorem \ref{theo:f-gan} of $\sup(I^*) \geq 0$ and $f(t)>0$ for $t\in [0,1)$. The corresponding $f$-GAN objective is
\begin{equation}   \label{eq:f-GAN-Wasserstein}
    \min_\nu \sup_{|T| \leq \alpha} \, \Bigl(\E_{\vx\sim \nu} [T(\vx)] -\E_{\vx\sim \mu} [T(\vx)]\Bigr).
\end{equation}
For $\alpha=1$, if we require $T$ to be continuous then the supremum part of \eqref{eq:f-GAN-Wasserstein} is precisely the total variation (also known as the {\em Radon metric}) between $\mu$ and $\nu$, which is closely related to the {\em Wasserstein distance} between $\mu$ and $\nu$.

\noindent\vspace{3mm}
{\bf Example 4:  \boldmath{$ f(t) = (t-1)\ln\frac{t}{t+1},~ t>0$} (the ``log\,D Trick'')}.

Here $f''(t) = \frac{3t+1}{t^2(t+1)^2}>0$ so $f$ is strictly convex. It satisfies all conditions of Theorem \ref{theo:f-gan}. The explicit expression for the convex dual $f^*$ is complicated to write down, however we do know the domain $I^*$ for $f^*$ is the range of $f'(t)$, which is $(-\infty,0)$. The $f$-GAN objective is
$$
\min_\nu \sup_{T<0} \, \Bigl(\E_{\vx\sim \nu} [T(\vx)] +\E_{\vx\sim \mu} [f^*(T(\vx))]\Bigr).
$$
By Theorem \ref{theo:sample-f-div-singular-2} and the fact $b^*=0$,
\begin{equation}  \label{eq:logD_trick}
   \sup_{T<0} ~ \Bigl(\E_{\vx\sim \mu} [T(\vx)] -\E_{\vx\sim \nu} [f^*(T(\vx))]\Bigr) =D_f(\mu\|\nu).
\end{equation}
Take $\tau = \frac{1}{2}(\mu+\nu)$ so $\mu,\nu\ll \tau$. Let $p = d\mu/d\tau$ and $q = d\nu/d\tau$. Observe that
\begin{align*}
   D_f(\mu\|\nu) &= \E_{\vx\sim\nu}\Bigl[f\Bigl(\frac{p(\vx)}{q(\vx)}\Bigr)\Bigr]
        = \E_{\vx\sim\nu}\Bigl[\Bigl(\frac{p(\vx)}{q(\vx)}-1\Bigr)\ln {\frac{p(\vx)}{p(\vx)+q(\vx)}}\Bigr]\\
        &= \E_{\vx\sim\mu}\Bigl[\ln\frac{p(\vx)}{p(\vx)+q(\vx)}\Bigr]
            - \E_{\vx\sim\nu}\Bigl[\ln\frac{p(\vx)}{p(\vx)+q(\vx)}\Bigr].
\end{align*}
Denote $D(x) = \frac{p(\vx)}{p(\vx)+q(\vx)}$. Then the outer minimization of the minimax problem is
\begin{equation}   \label{eq:f-GAN-logD}
    \min_\nu\, \Bigl(\E_{\vx\sim \mu} [\ln(D(\vx))] -\E_{\vx\sim \nu} [\ln(D(\vx))]\Bigr).
\end{equation}
This is precisely the ``\,$\log D$'' trick used in the original GAN paper \cite{Goodfellow14} for the vanilla GAN to address the saturation problem. Now we can see it is equivalent to the $f$-GAN with the above $f$. It is interesting to note that directly optimizing \eqref{eq:logD_trick} is hard because it is hard to find the explicit formula for $f^*$ in this case. Thus the vanilla GAN with the ``\,$\log D$ trick'' is an indirect way to realize this $f$-GAN.

\vspace{3mm}

To implement an $f$-GAN VDM we resort to the same approach as the vanilla GAN, using neural networks to approximate both $T(x)$ and $\nu$ to solve the minimax problem \eqref{eq:f-GAN}
$$
    \min_\nu \sup_{T} \, \Bigl(\E_{\vx\sim \nu} [T(\vx)] -\E_{\vx\sim \mu} [f^*(T(\vx))]\Bigr).
$$
We assume the critic function $T(x)$ comes from a neural network. \cite{NoCsTo16} proposes $T(x)= T_\omega(x) = g_f(S_\omega(x))$, where $S_\omega$ is a neural network with parameters $\omega$ taking input from $\R^n$ and $g_f:~\R \lra I^*$ is an {\em output activation function} to force the output from $V_\omega(x)$ onto the domain $I^*$ of $f^*$. For $\nu$ we again consider its approximation by probability distributions of the form $\nu_\theta = \gamma \circ G_\theta^{-1}$, where $\gamma$ is an initially chosen probability distribution on $\R^d$ (usually Gaussian, where $d$ may or may not be $n$), and $G_\theta$ is a neural network with parameters $\theta$, with input from $\R^d$ and output in $\R^n$. Under this model the $f$-GAN VMD minimax problem \eqref{eq:f-GAN} becomes
\begin{equation}  \label{eq:VDM-NN}
   \min_\theta \sup_{\omega} \,\Bigl(\E_{\vz\sim \gamma} [g_f(S_\omega(G_\theta(\vz)))] -\E_{\vx\sim \mu} [f^*(g_f(S_\omega(\vx)))]\Bigr).
\end{equation}

Like the vanilla GAN, since we do not have the explicit expression for the target distribution $\mu$, we shall approximate the expectations through sample averages. More specifically, let $\A$ be a minibatch of samples from the training data set $\X$ (a minibatch) and $\B$ be a minibatch of samples in $\R^d$ drawn from the distribution $\gamma$. Then we employ the approximations
\begin{align}
      \E_{\vz \sim \gamma} [g_f(S_\omega(G_\theta(\vz)))] & \approx \frac{1}{|\B|}\sum_{\vz\in\B} [g_f(S_\omega(G_\theta(\vz)))], \\
      \E_{\vx\sim \mu}[f^*(g_f(S_\omega(\vx)))] & \approx \frac{1}{|\A|}\sum_{\vx\in\A} f^*(g_f(S_\omega(\vx))).
\end{align}
The following algorithm for $f$-GAN VDM is almost a verbatim repeat of the Vanilla GAN Algorithm \cite{Goodfellow14} stated earlier:


\par\noindent\rule{\textwidth}{1.5pt}
{\bf VDM Algorithm} Minibatch stochastic gradient descent training of generative adversarial nets. Here $k\geq 1$ and $m$ are hyperparameters. \\
\noindent\rule{\textwidth}{1.0pt}

\noindent
for number of training iterations do \hfill\\
\hphantom{ab} for k steps do
\begin{itemize}
   \item Sample minibatch of $m$  samples $\{\vz_1, \dots, \vz_m\}$ in $\R^d$ from the distribution $\gamma$.
   \item Sample minibatch of $m$  samples $\{\vx_1, \dots, \vx_m\} \subset \X$ from the training set $\X$.
   \item Update $S_\omega$ by ascending its stochastic gradient with respect to $\omega$:
    $$
         \nabla_\omega \,\frac{1}{m} \sum_{i=1}^m \Bigl[g_f(S_\omega(G_\theta(\vz_i))) -f^*(g_f(S_\omega(\vx_i)))\Bigr].
    $$
\end{itemize}
\hphantom{ab} end for
\begin{itemize}
   \item Sample minibatch of $m$  samples $\{\vz_1, \dots, \vz_m\}$ in $\R^d$ from the distribution $\gamma$.
   \item Update the discriminator $G_\theta$ by descending its stochastic gradient with respect to $\theta$:
    $$
         \nabla_\theta \,\frac{1}{m} \sum_{i=1}^m g_f(S_\omega(G_\theta(\vz_i))).
    $$
\end{itemize}
end for

The gradient-based updates can use any standard gradient-based learning rule. \\
\noindent\rule{\textwidth}{0.5pt}

\section{Examples of Well-Known GANs}
\setcounter{equation}{0}

Since inception GANS have become one of the hottest research topics in machine learning. Many specially trained GANS tailor made for particular applications have been developed. Modifications and improvements have been proposed to address some of the shortcomings of vanilla GAN. Here we review some of the best known efforts in these directions.

\subsection{Wasserstein GAN (WGAN)}

Training a GAN can be difficult, which frequently encounters several failure modes. This has been a subject of many discussions. Some of the best known failure modes are:

\begin{description}
\item[{\boldmath$\cdot$} Vanishing Gradients]  This occurs quite often, especially when the discriminator is too good, which can stymie the improvement of the generator. With an optimal discriminator generator training can fail due to vanishing gradients, thus not providing enough information for the generator to improve.

\item[{\boldmath$\cdot$} Mode Collapse]   This refers to the phenomenon where the generator starts to produce the same output (or a small set of outputs) over and over again. If the discriminator gets stuck in a local minimum, then it's too easy for the next generator iteration to find the most plausible output for the current discriminator. Being stuck the discriminator never manages to learn its way out of the trap. As a result the generators rotate through a small set of output types.

\item[{\boldmath$\cdot$} Failure to Converge]  GANs frequently fail to converge, due to a number of factors (known and unknown).
\end{description}

The WGAN \cite{WGAN} makes a simple modification where it replaces the Jensen-Shannon divergence loss function in vanilla GAN with the Wasserstein distance, also known as the {\em Earth Mover (EM)} distance. Don't overlook the significance of this modification: It is one of the most important developments in the topic since the inception of GAN, as the use of EM distance effectively addresses some glaring shortcomings of divergence based GAN, allowing one to mitigate those common failure modes in the training of GANs.

Let $\mu,\nu$ be two probability distributions on $\R^n$ (or more generally any metric space). Denote by $\Pi(\mu,\nu)$ the set of all probability distributions $\pi(x,y)$  on $\R^n \times \R^n$ such that the marginals of $\pi$ are $\mu(x)$ and $\nu(y)$ respectively. Then the EM distance (Wasserstein-1 distance) between $\mu$ and $\nu$ is
$$
    W^1(\mu,\nu) := \min_{\pi \in \Pi(\mu,\nu)} \int_{\R^n \times \R^n} \|x-y\|\,d\pi(x,y)
                         = \min_{\pi \in \Pi(\mu,\nu)} \E_{(\vx,\vy)\sim \pi}[\|\vx-\vy\|].
$$
Intuitively $W^1(\mu,\nu)$ is called the earth mover distance because it denotes the least amount of work one needs to do to move mass $\mu$  to mass $\nu$.

In WGAN the objective is to minimize the loss function $W^1(\mu,\nu)$ as opposed to the loss function $D_f(\mu\|\nu)$ in $f$-GAN. The advantage of $W^1(\mu,\nu)$ is illustrated by \cite{WGAN} through the following example. Let $Z$ be the uniform distribution on $(0,1)$ in $\R$. Define $\mu = (0, Z)$ and $\nu_\theta = (\theta, Z)$ on $\R^2$. Then $\mu,\nu$ are singular distributions with disjoint support if $\theta \neq 0$. It is easy to check that
$$
       D_{JS}(\mu\|\nu_\theta) = \ln2, \quad \D_{KL}(\mu\|\nu_\theta) = \infty,
       \quad \mbox{and} \quad W^1(\mu,\nu_\theta)=|\theta|.
$$
However, visually for small $\theta >0$, even though $\mu$ and $\nu_\theta$ have disjoint support, they look very close. In fact if a GAN can approximate $\mu$ by $\nu_\theta$ for a very small $\theta$ we would be very happy with the result. But no matter how close $\theta>0$ is to 0 we will have $D_{JS}(\mu\|\nu_\theta) = \ln2$. If we train the vanilla GAN with the initial source distribution $\nu = \nu_\theta$ we would be stuck with a flat gradient so it will not converge. More generally,
let $\mu,\nu$ be probability measures such that $\mu \perp\nu$. Then we always have $D_{JS}(\mu\|\nu)= \ln2$. By Theorem \ref{theo:sample-f-div-singular-2} and \eqref{eq:f-div-singular-2}
$$
     \sup_{T}  \Bigl(\E_{\vx\sim \mu} [T(\vx)] -\E_{\vx\sim \nu} [f^*(T(\vx))]\Bigr) =\ln2.
$$
Thus gradient descend will fail to update $\nu$. In more practical setting, if our target distribution $\mu$ is a Gaussian mixture with well separated means, then starting with the standard Gaussian as initial source distribution will likely miss those Gaussian distributions in the mixture whose means are far away from 0, resulting in mode collapse and possibly failure to converge. Note that by the same Theorem \ref{theo:sample-f-div-singular-2} and \eqref{eq:f-div-singular-2}, things wouldn't improve by changing the convex function $f(x)$ in the $f$-GAN. As we seen from the examples, this pitfall can be avoided in WGAN.

The next question is how to evaluate $W^1(\mu,\nu)$ using only samples from the distributions, since we do not have explicit expression for the target distribution $\mu$. This is where Kantorovich-Rubenstein Duality \cite{villani-book} comes in, which states that
\begin{equation}  \label{eq:W1-dual}
    W^1(\mu,\nu) = \sup_{T\in {\rm Lip}_1(\R^n)} \Bigl(\E_{\vx\sim \mu} [T(\vx)] -\E_{\vx\sim \nu} [T(\vx)]\Bigr),
\end{equation}
where ${\rm Lip}_1(\R^n)$ denotes the set of all Lipschitz functions on $\R^n$ with Lipschitz constant 1. Here the critic $T(x)$ serves the role of a discriminator. With the duality WGAN solves the minimax problem
\begin{equation}  \label{eq:WGAN}
    \min_{\nu} W^1(\mu,\nu) = \min_\nu\sup_{T\in {\rm Lip}_1(\R^n)} \Bigl(\E_{\vx\sim \mu} [T(\vx)] -\E_{\vx\sim \nu} [T(\vx)]\Bigr).
\end{equation}
The rest follows the same script as vanilla GAN and $f$-GAN. We write $\nu = \gamma\circ G^{-1}$ where $\gamma$ is a prior source distribution (usually the standard normal) in $\R^d$ and $G$ maps $\R^d$ to $\R^n$. It follows that
$$
     \E_{\vx\sim \nu} [T(\vx)] = \E_{\vz\sim \gamma} [T(G(\vz))].
$$
Finally we approximate $T(x)$ and $G(z)$ by neural networks with parameters $\omega$ and $\theta$ respectively, $T(x) = T_\omega(x)$ and $G(z) = G_\theta(z)$. Stochastic gradient descend is used to train WGAN just like all other GANs. Indeed, replacing the JS-divergence with the EM distance $W^1$, the algorithm for vanilla GAN can be copied verbatim to become the algorithm for WGAN.

Actually {\em almost} verbatim. There is one last issue to be worked out, namely how does one enforce the condition $T_\omega(x) \in {\rm Lip}_1$? This is difficult. The authors have proposed a technique called {\em weight clipping}, where the parameter (weights) $\omega$ is artificially restricted to the region $\Omega:=\{\|\omega\|_\infty \leq 0.01\}$. In other words, all parameters in $\omega$ are clipped so they fall into the box $[-0.01,0.01]$. Obviously this is not the same as restricting the Lipschitz constant to 1. However, since $\Omega$ is compact, so will be $\{T_\omega: \omega\in \Omega\}$. This means the Lipschitz constant will be bounded by some $K>0$. The hope is that
$$
   \sup_{\omega\in\Omega} \Bigl(\E_{\vx\sim \mu} [T_\omega(\vx)] -\E_{\vx\sim \nu} [T_\omega(\vx)]\Bigr) \approx \sup_{T\in {\rm Lip}_K(\R^n)} \Bigl(\E_{\vx\sim \mu} [T(\vx)] -\E_{\vx\sim \nu} [T(\vx)]\Bigr),
$$
where the latter is just $K\cdot W^1(\mu,\nu)$.

Weight clipping may not be the best way to approximate the Lipschitz condition. Alternatives such as gradient restriction \cite{gulrajani2017improved} can be more effective. There might be other better ways.

\subsection{Deep Convolutional GAN (DCGAN)}

DCGAN refers to a set of architectural guidelines for GAN, developed in \cite{DCGAN}. Empirically the guidelines help GANs to attain more stable training and good performance. According to the paper, these guidelines are

\begin{itemize}
\item Replace any pooling layers with strided convolutions (discriminator) and fractional-strided
convolutions (generator).
\item Use batch normalization in both the generator and the discriminator.
\item Remove fully connected hidden layers for deeper architectures.
\item Use ReLU activation in generator for all layers except for the output, which uses Tanh.
\item Use LeakyReLU activation in the discriminator for all layers.
\end{itemize}
Here {\em strided convolution} refers to shifting the convolution window by more than 1 unit, which amounts to downsampling. Fractional strided convolution refers to shifting the convolution window by a fractional unit, say $1/2$ of a unit, which is often used for upsampling. This obviously cannot be done in the literal sense. To realize this we pad the input by zeros and then take the appropriate strided convolution. For example, suppose our input data is $X=[x_1, x_2, \dots, x_N]$ and the convolution window is $w=[w_1,w_2,w_3]$. For $1/2$ strided convolution we would first pad $X$ to become $\tilde X =[x_1,0,x_2,0, \dots,0,x_N]$ and then execute $\tilde X * w$. Nowadays, strided convolution is just one of the several ways for up and down sampling.

\subsection{Progressive Growing of GANs (PGGAN)}

Generating high resolution images from GANs is a very challenging problem. Progressive Growing of GANs developed in \cite{PGGAN} is a technique that addresses this challenge.

PGGAN actually refers to a training methodology for GANs. The
key idea is to grow both the generator and discriminator progressively: starting
from a low resolution image, one adds new layers that model increasingly fine details as
training progresses. Since low resolution images are much more stable and easier to train, the training is very stable in the beginning. Once training at a lower resolution is done, it gradually transit to training at a higher resolution. This process continues until the desired resolution is reached. In the paper \cite{PGGAN}, very high quality facial images have been generated by starting off the training at $4\times 4$ resolution and gradually increasing the resolution to $8\times 8$, $16\times 16$ etc, until it reaches $1024\times 1024$.

It would be interesting to provide a mathematical foundation for PGGAN. From earlier analysis there are pitfalls with GANs when the target distribution is singular, especially if it and the initial source distribution have disjoint supports. PGGAN may have provided an effective way to mitigate this problem.

\subsection{Cycle-Consistent Adversarial Networks (Cycle-GAN)}

Cycle-GAN is an image-to-image translation technique developed in \cite{Cycle-GAN}. Before this work the goal of image-to-image translation is to learn the mapping between an input image and an output image using a training set of aligned image pairs. However, often we may still want to do a similar task without the benefit of having a training set consisting of aligned image pairs . For example, while we have many paintings by Claude Monet, we don't have photographs of the scenes, and may wonder what those scenes would look like in a photo. We may wonder how a Monet painting of Mt. Everest would look like even though Claude Monet had never been to Mt. Everest. One way to achieve these tasks is through neural style transfer \cite{neural-style-transfer}, but this technique only transfers the style of a single painting to a target image. Cycle-GAN offers a different approach that allows for style transfer (tanslation) more broadly.

In a Cycle-GAN, we start with two training sets $\X$ and $\Y$. For example, $\X$ could be a corpus of Monet scenery paintings and $\Y$ could be a set of landscape photographs. The training objective of a Cycle-GAN is to transfer styles from $\X$ to $\Y$ and vice versa in a ``cycle-consistent'' way, as described below.

Precisely speaking, a Cycle-GAN consists of three components, each given by a tailored loss function. The first component is a GAN (vanilla GAN, but can also be any other GAN) that tries to generate the distribution of $\X$, with one notable deviation: instead of sampling initially from a random source distribution $\gamma$ such as a standard normal distribution, Cycle-GAN samples from the training data set $\Y$. Assume that $\X$ are samples drawn from the distribution $\mu$ and $\Y$ are samples drawn from the distribution $\nu_0$. The loss function for this component is
\begin{equation}  \label{eq:cycle-gan1}
    {\mathbf L}_{\rm gan1} (G_1, D_\mu) := \E_{\vx\sim \mu}[\log(D_\mu(\vx))] 
             + \E_{\vy\sim \nu_0}[\log(1-D_\mu(G_1(\vy)))]
\end{equation}
where $D_\mu$ is the discriminator network and $G_1$ is the generator network. Clearly this is the same loss used by the vanilla GAN, except the source distribution $\gamma$ is replaced by the distribution $\nu_0$ of $\Y$. The second component is the mirror of the first component, namely it is a GAN to learn the distribution 
$\nu_0$ of $\Y$ with the initial source distribution set as the distribution $\mu$ of $\X$. The corresponding loss function is thus
\begin{equation}  \label{eq:cycle-gan2}
{\mathbf L}_{\rm gan2} (G_2, D_{\nu_0}) := \E_{\vy\sim \nu_0}[\log(D_{\nu_0}(\vy))] 
+ \E_{\vx\sim \mu}[\log(1-D_{\nu_0}(G_2(\vy)))]
\end{equation}
where $D_{\nu_0}$ is the discriminator network and $G_2$ is the generator network. The third component of the Cycle-GAN is the ``cycle-consistent'' loss function given by
\begin{equation}  \label{eq:cycle-consistent}
{\mathbf L}_{\rm cycle} (G_1, G_2) := \E_{\vy\sim \nu_0}[\|G_2(G_1(\vy))-\vy\|_1] 
                      +\E_{\vx\sim \mu}[\|G_1(G_2(\vx))-\vx\|_1].
\end{equation}
The overall loss function is
\begin{equation}  \label{eq:cycle-loss}
{\mathbf L}^*(G_1,G_2,D_\mu,D_{\nu_0}) := {\mathbf L}_{\rm gan1}(G_1, D_\mu) 
       +{\mathbf L}_{\rm gan2} (G_2, D_{\nu_0}) + \lambda {\mathbf L}_{\rm cycle} (G_1, G_2),
\end{equation} 
where $\lambda>0$ is a parameter. Intuitively, $G_1$ translates a sample $\vy$ from the distribution $\nu_0$ into a sample from $\mu$, while $G_2$ translates a sample $\vx$ from the distribution $\mu$ into a sample from $\nu_0$. The loss function $ {\mathbf L}_{\rm cycle}$ encourages ``consistency'' in the sense that $G_2(G_1(\vy))$ is not too far off $\vy$ and $G_1(G_2(\vx))$ is not too far off $\vx$. Finally Cycle-GAN is trained by solving the minimax problem
\begin{equation}  \label{eq:cycle-minimax}
    \min_{G_1,G_2}\,\max_{D_\mu,D_{\nu_0}} \, {\mathbf L}^*(G_1,G_2,D_\mu,D_{\nu_0}).
\end{equation} 

\section{Alternative to GAN: Variational Autoencoder (VAE)}
\setcounter{equation}{0}

\begin{figure}  
	\includegraphics[width=0.7\textwidth]{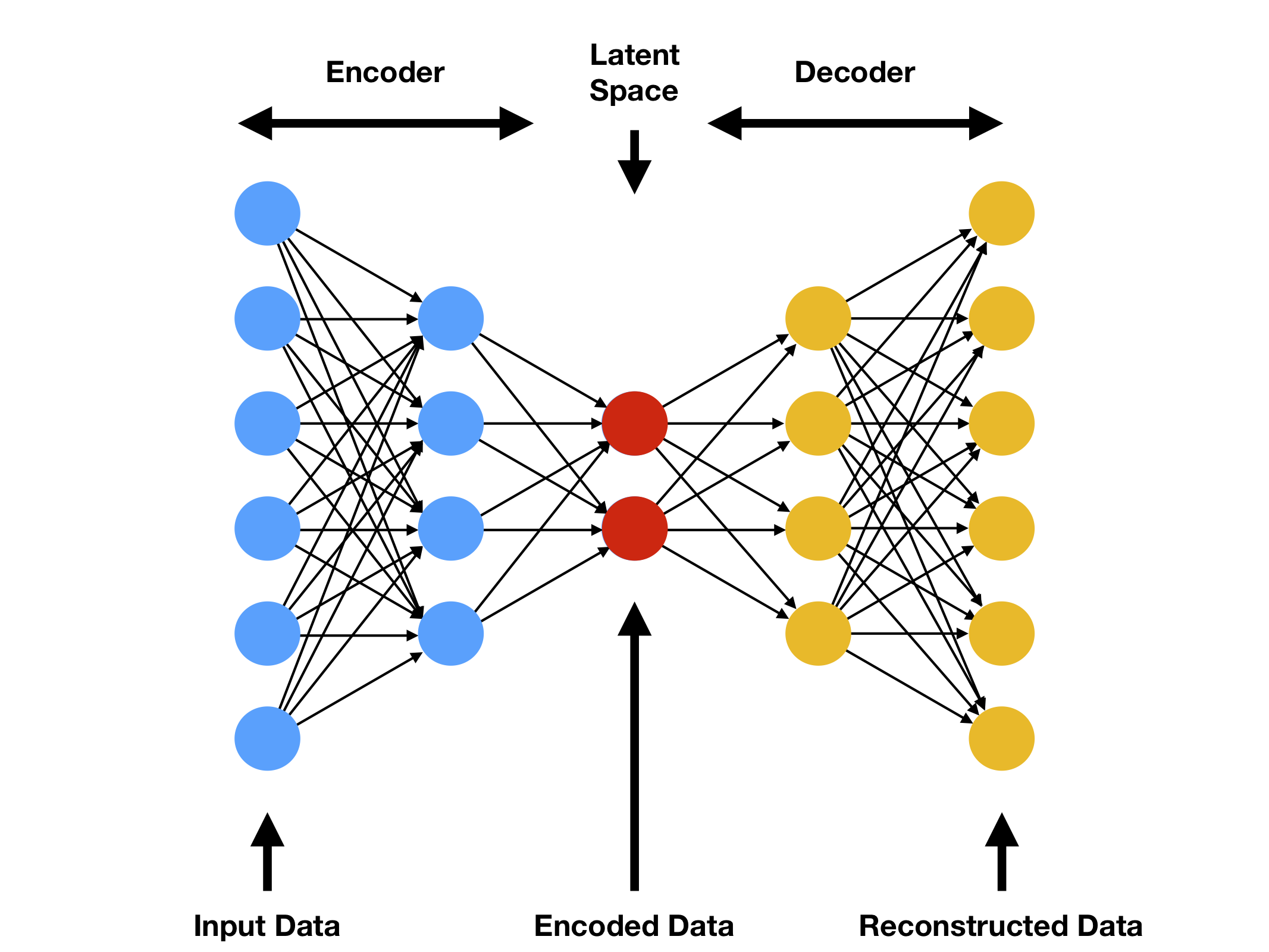}
	\caption{An autoencoder (courtsey of Jason Anderson and ThreeComp Inc.)}
	\label{fig:autoencoder}
\end{figure}

Variational Autoencoder (VAE) is an alternative generative model to GAN. To understand VAE one needs to first understand what is an {\em autoencoder}. In a nutshell an autoencoder consists of an encoder neural network $F$ that maps a dataset $\X \subset \R^n$ to $\R^d$, where $d$ is typically much smaller than $n$, together with a decoder neural network $H$ that ``decodes'' elements of $\R^d$ back to $\R^n$. In other words, it encodes the $n$-dimensional features of the dataset to the $d$-dimensional latents, along with a way to convert the latents back to the features. Autoencoders can be viewed as {\em data compressors} that compress a higher dimensional dataset to a much lower dimensional data set without losing too much information. For example, the MNIST dataset consists of images of size $28\times 28$, which is in $\R^{784}$. An autoencoder can easily compress it to a data set in $\R^{10}$ using only 10 latents without losing much information. A typical autoencoder has a ``bottleneck'' architecture, which is shown in Figure \ref{fig:autoencoder}.
%
The loss function for training an autoencoder is typically the mean square error (MSE) 
$$
   {\mathbf L}_{\rm AE} : = \frac{1}{N}\sum_{\vx\in\X} \|H(F(\vx)) -\vx\|^2
$$  
where $N$ is the size of $\X$. For binary data we may use the Bernoulli Cross Entropy (BCE) loss function. Furthermore, we may also add a regularization term to force desirable properties, e.g. a LASSO style $L^1$ loss to gain sparsity.

First developed in \cite{VAE}, VAEs are also neural networks having similar architectures as autoencoders, with stochasticity added into the networks.  Autoencoders are deterministic networks in the sense that output is completely determined by the input. To make generative models out of autoencoders we will need to add randomness to latents. In an autoencoder, input data $\vx$ are encoded to the latents $\vz=F(\vx)\in\R^d$, which are then decoded to $\hat\vx = H(F(\vz))$. A VAE deviates from an autoencoder in the following sense: the input $\vx$ is encoded into a diagonal Gaussian random variable $\vxi=\vxi(\vx)$ in $\R^d$ with mean $\vmu(\vx)$ and variance $\vsigma^2(\vx)$. Here $\vsigma^2(\vx)=[\sigma_1^2(\vx), \dots, \sigma_d^2(\vx)]^T \in\R^d$ and the variance is actually $\diag(\vsigma^2)$. Another way to look at this setup is that instead of having just one encoder $F$ in an autoencoder, a VAE neural network has two encoders $\vmu$ and $\vsigma^2$ for both the mean and the variance of the latent variable $\vxi$. As in an autoencoder, it also has a decoder $H$.  With randomness in place we now have a generative model.

\begin{figure}  
	\includegraphics[width=0.7\textwidth]{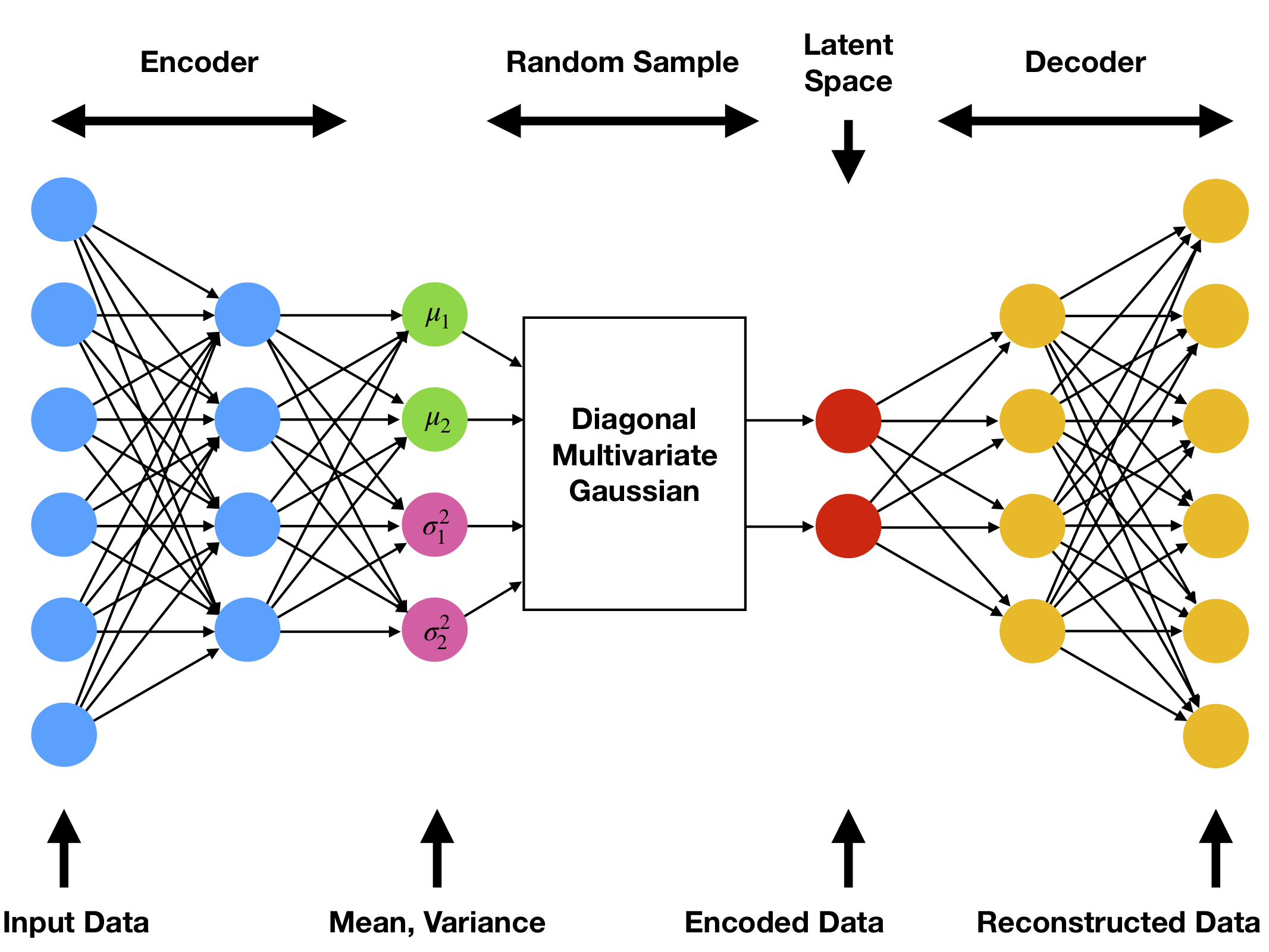}
	\caption{A variational autoencoder (courtesy of Jason Anderson and CompThree Inc.) }\label{fig:VAE}
\end{figure}

Of course we will need some constraints on $\vmu(\vx)$ and $\sigma^2(\vx)$. Here VAEs employ the following heuristics for training:
\begin{itemize}
	\item The decoder $H$ decodes the latent random variables $\vxi(\vx)$ to $\hat\vx$ that are close to $\vx$.
	\item The random variable $X=\vxi(\vx)$ with $\vx$ sampled uniformly from $\X$ is close to the standard normal distribution $N(0,1)$.
\end{itemize}
The heuristics will be realized through a loss function consisting of two components. The first component is simply the mean square error between $\hat\vx$ and $\vx$ given by
\begin{align}
{\mathbf L}_{1}(\vmu, \vsigma, G)  
 & = \frac{1}{N}\,\sum_{\vx\in\X} \E_{\vz\sim N(\vmu(\vx),\sigma(\vx)I_d)}[\|H(\vz) -\vx\|^2] \label{eq:vae1a}\\
           & = \frac{1}{N}\,\sum_{\vx\in\X} \E_{\vz\sim N(0,I_d)}[\|H(\vmu(\vx)+\sigma(\vx)\odot\vz) -\vx\|^2] \label{eq:vae1b}
\end{align} 
where $\odot$ denotes entry-wise product. Here going from \eqref{eq:vae1a} to \eqref{eq:vae1b} is a very useful technique called {\em re-parametrization}. The second component of the loss function is the KL-divergence (or other $f$-divergences) between $X$ and $N(0,I_d)$. For two Gaussian random variables their KL-divergence has an explicit expression, given by
\begin{equation} \label{eq:vae2}
     {\mathbf L}_{2}(\vmu, \vsigma)  = D_{KL}(X\|N(0,I_d))= \frac{1}{2N}\,\sum_{\vx\in\X} \sum_{i=1}^d \Bigl(\mu_i^2(\vx) +\sigma_i(\vx)^2-1+\ln(\sigma_i^2)\Bigr).
\end{equation}
The loss function for a VAE is thus
\begin{equation} \label{eq:vae}
{\mathbf L}_{\rm VAE}  = {\mathbf L}_{1} (\vmu, \vsigma, G) + \lambda \,{\mathbf L}_{2}(\vmu, \vsigma)
\end{equation}
where $\lambda >0$ is a parameter. To generate new data from a VAE, one inputs random samples $\vxi \sim N(0, I_d)$ into the decoder network $H$.  A typical VAE architecture is shown in Figure \ref{fig:VAE}.

This short introduction of VAE has just touched on the mathematical foundation of VAE. There have been a wealth of research focused on improving the performance of VAE and applications. We encourage interested readers to further study the subject by reading recent papers. 

\bibliographystyle{abbrv}
\bibliography{ref-gan}

\begin{thebibliography}{10}

\bibitem{AliSil65}
S.~M. Ali and S.~D. Silvey.
\newblock A general class of coefficients of divergence of one distribution
  from another.
\newblock {\em Journal of the Royal Statistical Society: Series B
  (Methodological)}, 28(1):131--142, 1966.

\bibitem{WGAN}
M.~Arjovsky, S.~Chintala, and L.~Bottou.
\newblock Wasserstein generative adversarial networks.
\newblock In {\em ICML}, 2017.

\bibitem{neural-style-transfer}
L.~A. Gatys, A.~S. Ecker, and M.~Bethge.
\newblock Image style transfer using convolutional neural networks.
\newblock In {\em Proceedings of the IEEE conference on computer vision and
  pattern recognition}, pages 2414--2423, 2016.

\bibitem{Goodfellow14}
I.~Goodfellow, J.~Pouget-Abadie, M.~Mirza, B.~Xu, D.~Warde-Farley, S.~Ozair,
  A.~Courville, and Y.~Bengio.
\newblock Generative adversarial nets.
\newblock In {\em NIPS}, 2014.

\bibitem{gulrajani2017improved}
I.~Gulrajani, F.~Ahmed, M.~Arjovsky, V.~Dumoulin, and A.~C. Courville.
\newblock Improved training of wasserstein gans.
\newblock In {\em Advances in neural information processing systems}, pages
  5767--5777, 2017.

\bibitem{PGGAN}
T.~Karras, T.~Aila, S.~Laine, and J.~Lehtinen.
\newblock Progressive growing of gans for improved quality, stability, and
  variation.
\newblock In {\em International Conference on Learning Representations}, 2018.

\bibitem{VAE}
D.~P. Kingma and M.~Welling.
\newblock Auto-encoding variational bayes.
\newblock {\em arXiv preprint arXiv:1312.6114}, 2013.

\bibitem{NgWaJo10}
X.~Nguyen, M.~J. Wainwright, and M.~I. Jordan.
\newblock Estimating divergence functionals and the likelihood ratio by convex
  risk minimization.
\newblock {\em IEEE Transactions on Information Theory}, 56(11):5847--5861,
  2010.

\bibitem{NoCsTo16}
S.~Nowozin, B.~Cseke, and R.~Tomioka.
\newblock f-gan: Training generative neural samplers using variational
  divergence minimization.
\newblock In {\em Advances in neural information processing systems}, pages
  271--279, 2016.

\bibitem{DCGAN}
A.~Radford, L.~Metz, and S.~Chintala.
\newblock Unsupervised representation learning with deep convolutional
  generative adversarial networks.
\newblock {\em arXiv preprint arXiv:1511.06434}, 2015.

\bibitem{villani-book}
C.~Villani.
\newblock {\em Optimal transport: old and new}, volume 338.
\newblock Springer Science \& Business Media, 2008.

\bibitem{Cycle-GAN}
J.-Y. Zhu, T.~Park, P.~Isola, and A.~A. Efros.
\newblock Unpaired image-to-image translation using cycle-consistent
  adversarial networks.
\newblock In {\em Proceedings of the IEEE international conference on computer
  vision}, pages 2223--2232, 2017.

\end{thebibliography}
\end{document}